\documentclass[runningheads]{llncs}
\usepackage[T1]{fontenc}
\usepackage{graphicx}
\usepackage{tabularx}
\usepackage{booktabs}
\usepackage{url}
\usepackage{mathtools}
\usepackage{comment}
\usepackage{amsmath,amssymb}
\usepackage{listings}

\usepackage{pgfplots}
\usepackage{multirow}

\usepackage{amsmath, amssymb}
\usepackage{tcolorbox}
\usepackage{enumitem}
\usepackage{verbatim}
\usepackage{caption}
\usepackage{subcaption}
\usepackage{tikz}
\usetikzlibrary{shapes.geometric, arrows.meta, positioning, fit, backgrounds}

\begin{document}
%
\title{DIALEVAL: Automated Type-Theoretic Evaluation of LLM Instruction Following}
%
%
\author{Nardine Basta\orcidID{0000-0001-5295-1375} \and
Dali Kaafar\orcidID{0000-0003-2714-0276}}
\authorrunning{N. Basta et al.}
%
\institute{Macquarie University, Australia\\
\email{nardine.basta@mq.edu.au}}
\maketitle              
\
\begin{abstract}
Evaluating instruction following in Large Language Models requires decomposing instructions into verifiable requirements and assessing satisfaction—tasks currently dependent on manual annotation and uniform criteria that do not align with human judgment patterns. We present DIALEVAL, a type-theoretic framework using dual LLM agents to automate instruction decomposition into typed predicates and implement type-specific satisfaction semantics. The framework enforces formal atomicity and independence constraints during automated extraction, then applies differentiated evaluation criteria—semantic equivalence for content predicates, exact precision for numerical predicates—mirroring empirically observed human assessment patterns. Extended to multi-turn dialogues through history-aware satisfaction functions, DIALEVAL enables evaluation in conversational contexts where single-turn methods fail. Validation demonstrates 90.38\% accuracy (26.45\% error reduction over baselines) and substantially stronger correlation with human judgment for complex instructions. 

\end{abstract}
\section{Introduction}

Reliable instruction following is fundamental for deploying Large Language Models in dialogue systems, where models must maintain conversational coherence while adhering to complex directives across multi-turn interactions. Current evaluation approaches face three critical limitations: (1) manual annotation of atomic requirements creates scalability bottlenecks and 20\%+ inter-annotator disagreement~\cite{zhou2023instruction,qin2023infobench}, (2) uniform evaluation criteria misalign with human judgment—humans accept semantic paraphrasing for content but demand exactness for numerical constraints~\cite{zhou2023instruction,wang2022}—causing recurring errors, and (3) single-turn paradigms cannot assess instruction adherence across conversational history and turn dependencies~\cite{ouyang2022training}. These limitations prevent systematic evaluation of dialogue systems for critical applications including task-oriented assistants and customer service agents.

We present DIALEVAL \footnote{The DIALEVAL implementation code and benchmark datasets are available at https://github.com/nardinebasta/DialEval}, a framework that reformulates instruction following evaluation as type-theoretic predicate satisfaction using a dual-agent architecture. Two specialized Claude-3.5-Sonnet agents separate instruction decomposition from evaluation. The first agent automatically extracts typed predicates (content, format, style, logical, numerical) from instructions with formal atomicity and independence constraints, eliminating manual annotation. The second agent performs type-specific satisfaction assessment using distinct criteria—lenient semantic matching for content, strict precision for numerical predicates—implemented through engineered prompts that mirror human evaluation patterns. For dialogue contexts, both agents incorporate conversational history through context-aware directives, enabling the evaluation of instruction adherence across multi-turn interactions while maintaining dialogue coherence.

Our contributions are: (1) \textbf{Automated type-theoretic evaluation framework}: Type-theoretic formalization of instructions as predicate sets with type-dependent satisfaction relations, enforcing atomicity and independence (no implicit satisfactions between predicates) through automated LLM-based decomposition that eliminates manual annotation. (2) \textbf{Type-specific evaluation semantics}:  Formalization of type-dependent satisfaction with distinct criteria per predicate type, aligning with human judgment while eliminating systematic errors from uniform evaluation. (3) \textbf{Context-aware dialogue evaluation}: Extension of instruction following assessment to multi-turn conversations through history-aware satisfaction functions, enabling systematic evaluation in dialogue systems where single-turn methods fail.

Empirical validation demonstrates 90.38\% accuracy compared to 86.92\% for state-of-the-art methods. For complex instructions, DIALEVAL exhibits substantially stronger alignment with human judgment (Pearson correlation 0.6517 versus 0.2612, $p < 0.001$). Application to multi-turn dialogues across GPT-3, GPT-4, DeepSeek, and Mixtral reveals three architectural patterns. First, all models demonstrate content predicate challenges (satisfaction scores 0.19-0.44) despite strong performance on style and logical predicates (scores $> 0.86$). Second, Mixtral exhibits architecture-specific format satisfaction weakness (0.40 versus 0.91-0.95 for other models). Third, dialogue initiative limitations persist across model scales, providing targeted insights for dialogue system development.

\section{Related Work}
\label{sec:related}


\textbf{Instruction Following Evaluation.} Early approaches relied on human assessment~\cite{ouyang2022training,zheng2023judging} with inter-annotator disagreement exceeding 20\%, while structured frameworks~\cite{taori2023stanford,dubois2023alpacafarm} improved consistency but remain difficult to scale. IFEval~\cite{zhou2023instruction} introduced verifiable instructions with deterministic rules but requires manual decomposition and applies only to specific types. INFOBENCH~\cite{qin2023infobench} provides detailed constraint categorization but depends entirely on manual annotation and lacks formal foundations. Recent work on constraint generation~\cite{liang2025fine} synthesizes training data through fine-grained constraint verification, complementing but not addressing automated evaluation.

\textbf{Automated Evaluation Frameworks.} LLM-as-evaluator approaches~\cite{zheng2023judging,liu2023geval,wang2023chatgpt,fu2023gptscore} apply uniform criteria across instruction types, failing to capture differential human assessment patterns. Critically, all existing methods operate on single-turn responses, unable to evaluate instruction adherence in multi-turn dialogue contexts. DIALEVAL addresses these limitations through automated type-specific evaluation with explicit dialogue support, providing the first formal framework for conversational instruction following assessment.

\section{Methodology}

DialEval employs two specialized agents implemented using Claude-3.5-Sonnet. The Instruction Analysis Agent ($\mathcal{A}_E$) decomposes instructions into typed predicates. The Evaluation Agent ($\mathcal{A}_S$) performs type-specific satisfaction assessment.
\subsection{Framework Architecture}

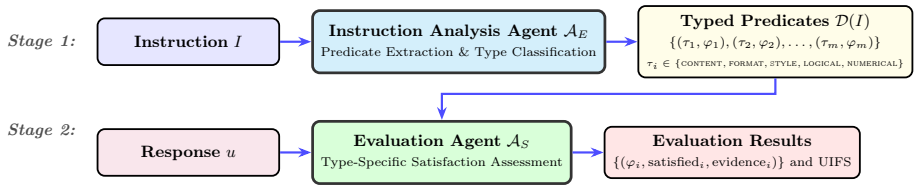
\begin{figure}[t]
\centering
\resizebox{\linewidth}{!}{
\begin{tikzpicture}[
    node distance=0.6cm,
    box/.style={rectangle, rounded corners, minimum width=3.5cm, minimum height=0.9cm, text centered, draw=black, line width=1pt, font=\small},
    agent/.style={rectangle, rounded corners, minimum width=3.5cm, minimum height=1.2cm, text centered, draw=black!80, line width=1.5pt, font=\small},
    arrow/.style={-{Stealth[length=2.5mm]}, line width=1.2pt},
    label/.style={font=\small\itshape, text=black!70}
]

\node[label] (stage1label) {\textbf{Stage 1:}};

\node[box, fill=blue!10, right=0.3cm of stage1label] (instruction) {
    \textbf{Instruction} $I$
};

\node[agent, fill=cyan!15, right=0.6cm of instruction] (agent1) {
    \begin{tabular}{c}
    \textbf{Instruction Analysis Agent} $\mathcal{A}_E$ \\
    \scriptsize Predicate Extraction \& Type Classification
    \end{tabular}
};

\node[box, fill=yellow!10, right=0.6cm of agent1, minimum height=1.3cm, minimum width=4.2cm] (predicates) {
    \begin{tabular}{c}
    \textbf{Typed Predicates} $\mathcal{D}(I)$ \\
    \scriptsize $\{(\tau_1, \varphi_1), (\tau_2, \varphi_2), \ldots, (\tau_m, \varphi_m)\}$ \\
    \tiny $\tau_i \in \{\textsc{content}, \textsc{format}, \textsc{style}, \textsc{logical}, \textsc{numerical}\}$
    \end{tabular}
};

\node[label, below=1.2cm of stage1label] (stage2label) {\textbf{Stage 2:}};

\node[box, fill=purple!10, below=1.2cm of instruction] (response) {
    \textbf{Response} $u$
};

\node[agent, fill=green!15, right=0.6cm of response] (agent2) {
    \begin{tabular}{c}
    \textbf{Evaluation Agent} $\mathcal{A}_S$ \\
    \scriptsize Type-Specific Satisfaction Assessment
    \end{tabular}
};

\node[box, fill=red!10, right=0.6cm of agent2, minimum width=4.2cm] (results) {
    \begin{tabular}{c}
    \textbf{Evaluation Results} \\
    \scriptsize $\{(\varphi_i, \mathrm{satisfied}_i, \mathrm{evidence}_i)\}$ and UIFS
    \end{tabular}
};

\draw[arrow, blue!70] (instruction.east) -- (agent1.west);
\draw[arrow, blue!70] (agent1.east) -- (predicates.west);

\draw[arrow, blue!70] (predicates.south) -- ++(0,-0.3) -| (agent2.north);
\draw[arrow, blue!70] (response.east) -- (agent2.west);

\draw[arrow, blue!70] (agent2.east) -- (results.west);

\end{tikzpicture}}
\caption{\footnotesize DIALEVAL dual-agent architecture. The Instruction Analysis Agent decomposes instructions into typed predicates. The Evaluation Agent performs type-specific satisfaction evaluation with the response, producing binary judgments and computing the Utterance-level Instruction Following Score (UIFS).}
\label{fig:architecture}
\vspace*{-0.6cm}
\end{figure}

The system architecture consists of two sequential processing stages as illustrated in Figure~\ref{fig:architecture}. The Instruction Analysis Agent receives an instruction $I$ and produces a structured set of typed predicates $\mathcal{D}(I) = \{(\tau_1, \varphi_1), \ldots, (\tau_m, \varphi_m)\}$, where each $\tau_i$ indicates the predicate type and $\varphi_i$ represents the satisfaction criterion. The Evaluation Agent then assesses a response $u$ against these predicates, yielding binary satisfaction judgments. The binary satisfaction judgments are aggregated into an Utterance-level Instruction Following Score (UIFS) representing the proportion of satisfied predicates (detailed in Section~\ref{sec:evaluation-agent}).

\textbf{Example.} Consider the instruction: ``Change the first person to the third person in the given sentence. The meaning should be kept, but you can paraphrase it or expand it in order to have a better pose.'' The Instruction Analysis Agent extracts four typed predicates: (1) $\varphi_1$ (\textsc{format}): ``Changes all first person pronouns to third person'', (2) $\varphi_2$ (\textsc{content}): ``Maintains the same core meaning'', (3) $\varphi_3$ (\textsc{style}): ``Includes acceptable paraphrasing or expansion'' with dependency on $\varphi_2$, and (4) $\varphi_4$ (\textsc{style}): ``Maintains better prose/flow'' with dependencies on $\varphi_2$ and $\varphi_3$.

Given the input ``We were recently able to increase the amount of stock we hold with the same supplier thereby reducing our risk,'' GPT-4 generates: ``They were recently able to increase the amount of stock they hold with the same supplier, thereby reducing their risk.'' The Evaluation Agent assesses this response using type-specific semantics: $\varphi_1$ and $\varphi_2$ are satisfied (correct pronoun conversion and meaning preservation), while $\varphi_3$ and $\varphi_4$ fail (no paraphrasing or prose improvement). This yields UIFS = 0.50, with type-specific scores revealing perfect \textsc{format}/\textsc{content} adherence (1.0) but complete \textsc{style} failure (0.0).

\subsection{Theoretical Foundation}

We represent evaluation contexts as $\mathcal{D} = (I, \mathcal{U})$, where $I$ denotes the instruction and $\mathcal{U} = \{u_1, \ldots, u_n\}$ represents response utterances. We classify predicates into five types: $\mathcal{T} = \{\textsc{content}, \textsc{format}, \textsc{style}, \textsc{logical}, \textsc{numerical}\}$. This typing system is derived from systematic analysis of instruction patterns in established benchmarks and empirical studies~\cite{zhou2023instruction,qin2023infobench}. Each predicate is formalized as:
\begin{equation}
\varphi_i = (\tau_i, \textit{pred}_i)
\end{equation}
where $\tau_i \in \mathcal{T}$ indicates the predicate type, and $\textit{pred}_i: u_j \to \{\top, \bot\}$ is a satisfaction function mapping utterances to true ($\top$) or false ($\bot$).

The satisfaction relation is type-dependent, written as $u \models_\tau \varphi$, meaning utterance $u$ satisfies predicate $\varphi$ under type $\tau$'s evaluation semantics:
\begin{equation}
u \models_\tau \varphi \iff \textit{eval}_\tau(\varphi, u) = \top
\end{equation}
where $\textit{eval}_\tau$ implements type-specific evaluation criteria. This type-dependence reflects empirical findings that human evaluators apply fundamentally different assessment strategies for different constraint categories~\cite{qin2023infobench,wang2022}. For instance, humans tolerate semantic paraphrasing for content requirements but demand exact matching for numerical constraints~\cite{zhou2023instruction}. For the predicate ``include the value 42,'' the response ``approximately 42'' satisfies under \textsc{content} semantics (semantic equivalence accepted) but fails under \textsc{numerical} semantics (exactness required). Table~\ref{tab:type-semantics} summarizes these semantic distinctions.

\begin{table}[t]
\centering
\caption{\footnotesize Type-dependent satisfaction semantics}
\label{tab:type-semantics}
\footnotesize
\resizebox{0.9\columnwidth}{!}{
\begin{tabular}{p{0.15\linewidth}p{0.9\linewidth}}
\hline
\textbf{Type} & \textbf{Evaluation Characteristics} \\
\hline
\textsc{content} & Semantic equivalence; flexible phrasing; focuses on information presence \\
\textsc{format} & Structural conformance; functional variations acceptable \\
\textsc{style} & Holistic impression; overall tone assessment \\
\textsc{logical} & Reasoning structure; validates key logical connections \\
\textsc{numerical} & Strict precision; exact matching; no approximations \\
\hline
\end{tabular}}
\end{table}

\subsection{Instruction Analysis Agent}

The analysis agent $\mathcal{A}_E$ decomposes instructions through a two-stage process. First, predicate extraction maps instruction $I$ to atomic requirements:

\begin{equation}
\mathcal{D}(I) = \{(\tau_1, \varphi_1), (\tau_2, \varphi_2), \ldots, (\tau_m, \varphi_m)\}
\end{equation}
Predicates must satisfy two formal criteria. (1) Semantic atomicity ensures each predicate represents an indivisible requirement.
(2) Operational independence ensures predicates do not implicitly satisfy one another.
These theoretical criteria are implemented through specific directives in the analyzer prompt. Figure~\ref{fig:analyzer-prompt} shows how atomicity is enforced through instructions that each requirement should ``represent a single, indivisible task'' and ``can be verified independently.''
\begin{figure}[t]
\centering

\begin{subfigure}[b]{0.47\textwidth}
\centering
\fcolorbox{black!30}{blue!5}{%
\begin{minipage}{1\textwidth}
\vspace{1pt}
\scriptsize
\textbf{Instruction Analysis Agent Directives}
\vspace{1pt}

\hrule
\vspace{2pt}

\scriptsize
\textbf{1. Extract BOTH explicit \& implicit requirements}
\begin{itemize}[leftmargin=8pt, itemsep=0pt, topsep=1pt]
    \item Each requirement independently verifiable
    \item Each represents single, indivisible task
    \item Avoid excessive decomposition (5-7 core requirements)
\end{itemize}

\vspace{1pt}
\textbf{2. Classify each predicate into ONE type:}
\begin{itemize}[leftmargin=8pt, itemsep=0pt, topsep=1pt]
    \item \textbf{content}: Information/content requirements \\
    \textit{(Does response include required content?)}
    \item \textbf{format}: Structural/formatting requirements \\
    \textit{(Does response follow required structure?)}
    \item \textbf{style}: Writing style/tone requirements \\
    \textit{(Does response match required tone/style?)}
    \item \textbf{logical}: Reasoning/logic requirements \\
    \textit{(Does response demonstrate correct logic?)}
    \item \textbf{numerical}: Mathematical/quantitative requirements \\
    \textit{(Are numerical values correct?)}
\end{itemize}

\vspace{1pt}
\end{minipage}%
}
\caption{\footnotesize Analyzer prompt 
}
\label{fig:analyzer-prompt}
\end{subfigure}
\hfill
\begin{subfigure}[b]{0.515\textwidth}
\centering
\fcolorbox{black!30}{green!5}{%
\begin{minipage}{1\textwidth}
\vspace{1pt}
\scriptsize
\textbf{Evaluation Agent Type-Specific Criteria}
\vspace{1pt}

\hrule
\vspace{2pt}

\scriptsize
\textbf{\textsc{content} Predicates:}
\begin{itemize}[leftmargin=8pt, itemsep=0pt, topsep=1pt]
    \item Verify required information substantially present
    \item Allow different phrasings conveying same information
    \item \textit{BE GENEROUS} in assessing content satisfaction
\end{itemize}

\vspace{1pt}
\textbf{\textsc{format} Predicates:}
\begin{itemize}[leftmargin=8pt, itemsep=0pt, topsep=1pt]
    \item Check adherence to structural requirements
    \item Recognize when slight variations serve functional purpose
    \item \textit{DO NOT OVERWEIGHT} formatting vs. content
\end{itemize}

\vspace{1pt}
\textbf{\textsc{numerical} Predicates:}
\begin{itemize}[leftmargin=8pt, itemsep=0pt, topsep=1pt]
    \item Verify mathematical accuracy
    \item Each numerical requirement must be strictly correct
    \item \textit{NO APPROXIMATIONS} acceptable
\end{itemize}

\vspace{1pt}
\textit{Provide binary satisfaction (true/false) with specific evidence from response text}

\vspace{1pt}
\end{minipage}%
}
\caption{\footnotesize Evaluator prompt 
}
\label{fig:evaluator-prompt}
\end{subfigure}

\caption{\footnotesize Core DIALEVAL prompts for instruction analysis (a) and evaluation (b).} 
\label{fig:core-prompts}
\vspace*{-0.6cm}
\end{figure}

Second, the type classification assigns each predicate to one of five types through prompt-based inference over explicit definitions and exemplars. For instance, \textsc{content} constraints are specified as ``information/content requirements (Does the response include required content?)'' while \textsc{format} constraints are ``structural/formatting requirements (Does the response follow the required structure?).'' The classification function $\mathcal{T}(\varphi_i) : \varphi_i \mapsto \tau \in \mathcal{T}$ selects the type $\tau$ with highest semantic similarity to predicate $\varphi_i$ based on these definitions and examples in knowledge base of type definitions and examples $K$. The output is a JSON object containing typed predicates, enabling systematic evaluation by the Evaluation agent.

\subsection{Evaluation Agent}
\label{sec:evaluation-agent}

The evaluation agent $\mathcal{A}_S$ implements type-specific assessment strategies through specialized prompt templates. The evaluation function is defined as:

\begin{equation}
\textit{eval}(\varphi_i, u_j) = \mathcal{A}_S(\pi_{\tau_i}(\varphi_i, u_j))
\end{equation}
where $\pi_{\tau_i}$ is the type-specific prompt template for type $\tau_i$. Each evaluation yields a binary satisfaction result with supporting evidence:

\begin{equation}
\textit{eval}(\varphi_i, u_j) = [\textit{satisfied}, \textit{evidence}]
\end{equation}
where $\textit{satisfied} \in \{\top, \bot\}$ and $\textit{evidence}$ is textual justification extracted from $u_j$.

The type-specific prompt templates $\pi_{\tau_i}$ are implemented through distinctive evaluation criteria for each predicate type in the evaluator prompt. Figure~\ref{fig:evaluator-prompt} shows how different types receive different evaluation directives.

For \textsc{content} predicates, the prompt instructs the agent to ``verify required information is substantially present'' and ``allow for different phrasings that convey the same information.'' For \textsc{format} predicates, it directs checking ``adherence to structural requirements'' while recognizing when ``slight variations still serve the functional purpose.'' For \textsc{numerical} predicates, strict precision is enforced: ``each numerical requirement must be strictly and precisely correct.''

The binary outcome is implemented through the prompt directive to provide ``binary satisfaction (true/false) that aligns with how a human would judge'' along with ``specific evidence from the response text.'' The prompt emphasizes that evaluation should ``match your judgments to how an average human would evaluate the same response,'' ensuring human-aligned assessment.

The utterance-level instruction following score (UIFS) quantifies adherence as the proportion of satisfied predicates:
\begin{equation}\label{eq:uifs}
\textit{UIFS}(I, u_j) = \frac{|\mathcal{S}_j|}{|\mathcal{D}(I)|}
\end{equation}
where $\mathcal{S}_j = \{\varphi_i \in \mathcal{D}(I) \mid \textit{eval}(\varphi_i, u_j).\textit{satisfied} = \top\}$ is the set of satisfied predicates and $|\mathcal{D}(I)|$ is the total number of predicates. 

\subsection{Dialogue-Specific Extensions}
For multi-turn dialogues, we extend the evaluation framework to incorporate conversational context through history-aware prompts:
\begin{equation}
\textit{eval}_{\textit{dialogue}}(\varphi_i, u_j, h_j) = \mathcal{A}_S(\pi_{\tau_i}(\varphi_i, u_j, h_j))
\end{equation}
where $h_j$ represents dialogue history up to utterance $u_j$, including the most recent message from the conversation partner. Both agents receive dialogue-specific directives: the analyzer considers conversational dynamics when extracting predicates (Figure~\ref{fig:dialogue-analyzer}), while the evaluator incorporates dialogue history into satisfaction assessment (Figure~\ref{fig:dialogue-evaluator}).

\begin{figure}[t]
\centering

\begin{subfigure}[b]{0.49\textwidth}
\centering
\fcolorbox{black!30}{cyan!5}{%
\begin{minipage}{1\textwidth}
\vspace{1pt}
\scriptsize
\textbf{Dialogue Analysis Extension}
\vspace{1pt}

\hrule
\vspace{2pt}

\scriptsize
\textit{These instructions evaluate responses in a DIALOGUE context. Consider the following when extracting predicates:}
\vspace{2pt}

\textbf{1. Conversational dynamics:}
\begin{itemize}[leftmargin=8pt, itemsep=0pt, topsep=1pt]
    \item Each response relates to preceding messages
    \item Instructions may apply across multiple turns
\end{itemize}

\vspace{1pt}
\textbf{2. Dialogue coherence:}
\begin{itemize}[leftmargin=8pt, itemsep=0pt, topsep=1pt]
    \item Predicates must account for maintaining conversation flow
    \item Balance instruction following with natural dialogue
\end{itemize}

\vspace{1pt}
\textbf{3. Context-sensitive behavior:}
\begin{itemize}[leftmargin=8pt, itemsep=0pt, topsep=1pt]
    \item Requirements interpreted based on dialogue context
    \item Previous exchanges inform predicate interpretation
\end{itemize}

\vspace{1pt}
\textbf{4. Turn-by-turn dependencies:}
\begin{itemize}[leftmargin=8pt, itemsep=0pt, topsep=1pt]
    \item Predicates may build upon prior turns
    \item Extract requirements considering sequential nature
\end{itemize}

\vspace{1pt}
\end{minipage}%
}
\caption{\footnotesize Analyzer dialogue context directives}
\label{fig:dialogue-analyzer}
\end{subfigure}
\hfill
\begin{subfigure}[b]{0.49\textwidth}
\centering
\fcolorbox{black!30}{orange!5}{%
\begin{minipage}{1\textwidth}
\vspace{1pt}
\scriptsize
\textbf{Dialogue Evaluation Extension}
\vspace{1pt}

\hrule
\vspace{2pt}

\scriptsize
\textit{Evaluate the response in the context of the MOST RECENT utterance from the other party in the dialogue. Consider:}
\vspace{2pt}

\textbf{1. Content addressing:}
\begin{itemize}[leftmargin=8pt, itemsep=0pt, topsep=1pt]
    \item Does response address questions/content from partner?
    \item Is information relevant to conversational context?
\end{itemize}

\vspace{1pt}
\textbf{2. Conversational flow:}
\begin{itemize}[leftmargin=8pt, itemsep=0pt, topsep=1pt]
    \item Does response maintain natural dialogue progression?
    \item Is turn-taking appropriate and coherent?
\end{itemize}

\vspace{1pt}
\textbf{3. History awareness:}
\begin{itemize}[leftmargin=8pt, itemsep=0pt, topsep=1pt]
    \item Does response demonstrate understanding of context?
    \item Are references to prior turns appropriate?
\end{itemize}

\vspace{1pt}
\textbf{4. Instruction-dialogue balance:}
\begin{itemize}[leftmargin=8pt, itemsep=0pt, topsep=1pt]
    \item Balance instruction following with conversational naturalness
    \item Assess if predicate satisfaction maintains dialogue quality
\end{itemize}

\vspace{1pt}
\end{minipage}%
}
\caption{\footnotesize Evaluator dialogue context directives}
\label{fig:dialogue-evaluator}
\end{subfigure}

\caption{\footnotesize DIALEVAL dialogue-specific extensions incorporating dialogue history for instruction analysis (a) and evaluation (b).} 
\label{fig:dialogue-prompts}
\vspace*{-0.7cm}
\end{figure}

The dialogue-level instruction following score (DIFS) aggregates utterance-level scores across the conversation:
\begin{equation}
\textit{DIFS}(I, \mathcal{U}) = \frac{1}{|\mathcal{U}|} \sum_{j=1}^{|\mathcal{U}|} \textit{UIFS}(I, u_j)
\end{equation}
where $|\mathcal{U}|$ is the number of utterances in the dialogue. This formulation enables assessment of instruction following consistency across multi-turn interactions.


\section{Experimental Setup}
We evaluate DIALEVAL through two complementary assessments: (1) validation against human annotation on standard benchmarks, and (2) evaluation of instruction following in multi-turn dialogues. 

\textbf{Datasets. }
(1)\textit{INFOBENCH Validation Dataset. }For validation against human judgment, we use INFOBENCH~\cite{qin2023infobench}, comprising 500 instructions with manually decomposed requirements across Easy Set (252 instructions, 71 domains) and Hard Set (248 instructions, 72 diverse domains). We utilize all 50 instances with existing expert annotations (25 Easy, 25 Hard). Each instance contains an instruction prompt, manually decomposed atomic instructions and responses from four LLMs (GPT-3.5-Turbo, GPT-4, Claude-v1, Vicuna-13B), yielding 200 instruction-response pairs (50 instructions × 4 models). Three independent annotators provide binary satisfaction judgments for each pair.

(2)\textit{BotWars Dialogue Dataset. }For dialogue evaluation, we use the BotWars dataset \cite{botwarspaper}, containing adversarial conversations between simulated scammers and potential victims. We randomly sample 80 conversations (20 per model) generated by GPT-3, GPT-4, DeepSeek, and Mixtral, with lengths ranging from 6 to 50 utterances (mean: 33). This dataset enables testing instruction following in dynamic conversations.

\textbf{Evaluation Benchmarks. }(1)\textit{Single-Turn Validation. }We compare DIALEVAL vs. the INFOBENCH GPT-based evaluator~\cite{qin2023infobench}, the state-of-the-art automated evaluation system achieving 86.92\% agreement with human judgment. 

(2)\textit{Dialogue Evaluation. }No benchmark exists for automated instruction following evaluation in multi-turn dialogues. We validate DIALEVAL through convergent validation by examining whether predicate-type-specific scores align with established architectural characteristics~\cite{hendrycks2020measuring}. 

\textbf{Evaluation Metrics. }
(1)\textit{Accuracy.} We measure prompt-level accuracy by comparing aggregate satisfaction scores. Since DIALEVAL generates different predicate decompositions than manual annotations, we define accuracy as:
\begin{equation}
\text{Accuracy} = 1 - \frac{1}{N} \sum_{i=1}^{N} |S_{\text{human}}(i) - S_{\text{DIALEVAL}}(i)|
\end{equation}
where $N$ is the number of prompt-response pairs, $S_{\text{human}}(i)$ is the human-annotated satisfaction score (proportion of satisfied predicates), and $S_{\text{DIALEVAL}}(i)$ is DIALEVAL's UIFS score (Equation~\ref{eq:uifs}).

(2)\textit{Correlation Analysis.} We assess statistical alignment between DIALEVAL and human judgments using Pearson's correlation coefficient ($r$) for linear relationships and Spearman's rank correlation ($\rho$) for monotonic relationships. Both metrics range from -1 to 1, with values closer to 1 indicating stronger positive correlation. We report correlation values with statistical significance calculated using two-tailed tests ($\alpha = 0.05$).

(3)\textit{Type-Specific Performance.} For dialogue evaluation, we calculate predicate-type-specific scores to identify performance variations across the five types: content, format, style, logical, and numerical. For each type $\tau \in \mathcal{T}$, the type-specific satisfaction score is:
\begin{equation}
\text{Score}_\tau(I, \mathcal{U}) = \frac{|\{\varphi_i \in \mathcal{D}(I) \mid \tau_i = \tau \wedge u_j \models_\tau \varphi_i\}|}{|\{\varphi_i \in \mathcal{D}(I) \mid \tau_i = \tau\}|}
\end{equation}
where $\mathcal{D}(I)$ is the set of typed predicates extracted from instruction $I$, $\tau_i$ is the type of predicate $\varphi_i$, and $u_j \models_\tau \varphi_i$ denotes satisfaction under type-specific semantics. This analysis reveals architectural strengths and weaknesses of different LLMs across predicate categories.

\section{Evaluation and Results}

\subsection{Validation Against Human Evaluation}

\begin{figure*}[h]
\vspace*{-0.3cm} 
    \centering
    \begin{subfigure}[b]{0.49\textwidth}
        \centering
\resizebox{1\linewidth}{!}{
\begin{tikzpicture}
\begin{axis}[
   width=12cm,
   height=7.5cm,
   xlabel={\textbf{Evaluation Set}},
   ylabel={\textbf{Accuracy (\%)}},
   symbolic x coords={Hard Set, Easy Set, Combined},
   xtick=data,
   ytick={70,75,80,85,90,95,100},
   ymin=70,
   ymax=100,
   legend style={at={(0.98,0.99)}, anchor=north east, draw=gray!40, fill=white!95, opacity=0.9},
   ymajorgrids=true,
   grid style={dotted, gray!50},
   axis background/.style={fill=gray!5},
   nodes near coords={\pgfmathprintnumber[precision=2]{\pgfplotspointmeta}\%},
   every node near coord/.append style={font=\small\bfseries, yshift=0.35cm},
   bar width=0.95cm,
   ybar=0.2cm,
   enlarge x limits={0.2},
]
\addplot[fill=teal!80, draw=teal!60, ultra thick] coordinates {(Hard Set, 89.52) (Easy Set, 91.23) (Combined, 90.38)};
\addplot[fill=orange!80, draw=orange!60, ultra thick] coordinates {(Hard Set, 84.34) (Easy Set, 89.50) (Combined, 86.92)};
\legend{\textbf{DIALEVAL}, \textbf{INFOBENCH}}
\end{axis}
\end{tikzpicture}}
\caption{\footnotesize Comparison of DIALEVAL and INFOBENCH GPT Evaluator accuracy}
\label{fig:accuracy_comparison}
    \end{subfigure}
    \hfill
    \begin{subfigure}[b]{0.49\textwidth}
        \centering
\resizebox{1\linewidth}{!}{
\begin{tikzpicture}
\begin{axis}[
   width=12cm,
   height=7.5 cm,
   xlabel={Dataset},
   ylabel={Correlation Coefficient},
   symbolic x coords={DIALEVAL (Hard), GPT (Hard), DIALEVAL (Easy), GPT (Easy)},
   xtick=data,
   ymin=0,
   ymax=0.8,
   legend style={at={(0.95,0.95)}, anchor=north east, draw=gray!30, fill=white!90, opacity=0.9},
   ymajorgrids=true,
   grid style=dashed,
   nodes near coords,
   every node near coord/.append style={font=\small},
   ybar=0.1cm,
   bar width=0.7cm
]
\addplot[fill=teal!80, draw=teal!80!black] coordinates {(DIALEVAL (Hard), 0.6517) (GPT (Hard), 0.2612) (DIALEVAL (Easy), 0.3610) (GPT (Easy), 0.0957)};
\addplot[fill=orange!70, draw=orange!80!black] coordinates {(DIALEVAL (Hard), 0.5767) (GPT (Hard), 0.2835) (DIALEVAL (Easy), 0.3016) (GPT (Easy), 0.2072)};
\legend{Pearson, Spearman}
\end{axis}
\end{tikzpicture}}
\caption{\footnotesize Correlation coefficients of evaluation systems with human majority votes}
\label{fig:correlation_comparison}
    \end{subfigure}
    \vfill

    \caption{\small DIALEVAL validation against human evaluation.}
    \label{fig:results}
    \vspace*{-0.8cm}
\end{figure*}
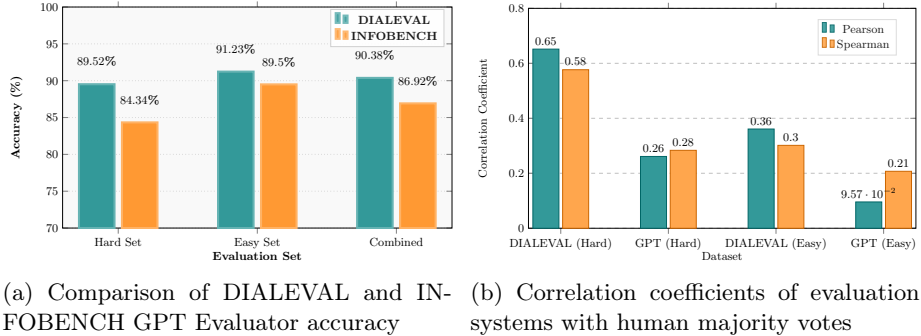

DIALEVAL achieves 90.38\% overall accuracy against human majority voting compared to 86.92\% for the INFOBENCH GPT-based evaluator ($p < 0.01$, paired t-test), representing 26.45\% error reduction (Figure~\ref{fig:accuracy_comparison}). 

\textbf{Complexity-Based Analysis.} Performance differences are most pronounced on complex instructions. While both systems perform comparably on the Easy Set (DIALEVAL: 91.23\%, INFOBENCH: 89.50\%), DIALEVAL demonstrates superior performance on the Hard Set (89.52\% versus 84.34\%), indicating that type-specific evaluation semantics provide advantages for multi-faceted instructions requiring careful predicate management.

Correlation analysis further confirms this advantage (Figure~\ref{fig:correlation_comparison}). For the Hard Set, DIALEVAL achieves Pearson correlation of 0.6517 ($p < 0.0001$) with human judgment versus INFOBENCH's 0.2612 ($p = 0.0087$), a statistically significant improvement. This demonstrates that type-dependent satisfaction semantics—semantic equivalence for content predicates, exact precision for numerical predicates—effectively capture human assessment patterns for complex instructions.

\textbf{Model-Specific Analysis. }DIALEVAL demonstrates consistent performance across all evaluated models (Table~\ref{table:model_accuracy}). Notable patterns include high correlation with Claude-v1 ($r = 0.8179$, $p < 0.0001$ for Easy Set), while the INFOBENCH evaluator shows higher accuracy on GPT models, reflecting architecture-specific bias. DIALEVAL's advantage is most pronounced for open-source models: on Vicuna-13b Hard Set, DIALEVAL achieves 89.01\% accuracy versus INFOBENCH's 75.45\% (55.25\% error reduction). This demonstrates DIALEVAL's capability to evaluate diverse architectures through type-theoretic satisfaction criteria rather than implicit pattern matching.

\begin{table}[h]
\vspace*{-0.8cm}
\centering
\caption{\footnotesize Model-specific accuracy comparison}
\label{table:model_accuracy}
\resizebox{0.6\columnwidth}{!}{
\begin{tabular}{lcccc}
\toprule
\multirow{2}{*}{Model} & \multicolumn{2}{c}{Easy Set} & \multicolumn{2}{c}{Hard Set} \\
\cmidrule(lr){2-3} \cmidrule(lr){4-5}
& DIALEVAL & INFOBENCH & DIALEVAL & INFOBENCH \\
\midrule
GPT-4 & 95.63\% & 94.00\% & 87.66\% & 85.73\% \\
Claude-v1 & 94.55\% & 86.33\% & 92.65\% & 86.06\% \\
GPT-3.5-turbo & 90.00\% & 94.67\% & 88.76\% & 90.10\% \\
Vicuna-13b & 84.92\% & 83.00\% & 89.01\% & 75.45\% \\
\bottomrule
\end{tabular}}
\vspace*{-0.4cm}
\end{table}

\textbf{Error Pattern Analysis. }DIALEVAL exhibits more balanced error distribution than INFOBENCH: false positive to false negative ratio of 1.23:1 versus 2.17:1 for the Hard Set, indicating INFOBENCH's tendency toward over-optimistic assessment. Critically, 68\% of DIALEVAL's disagreements with human majority voting occurred in boundary cases where annotators themselves disagreed, suggesting these divergences reflect genuinely ambiguous cases rather than evaluation failures. This demonstrates that DIALEVAL's type-theoretic approach effectively captures human judgment patterns, including inherent ambiguity in complex instructions.


\begin{table*}[ht]
\vspace*{-0.7cm}
\captionsetup{font=footnotesize}
    \parbox{.48\linewidth}{
        \centering
    \caption{\footnotesize{Distribution of error types across evaluation systems}}
    \label{table:error_types}
    \resizebox{0.48\columnwidth}{!}{
    \begin{tabular}{lcc}
\toprule
Error Type & DIALEVAL & INFOBENCH\\
\midrule
False Positives & 6.8\% & 9.4\% \\
False Negatives & 5.5\% & 4.3\% \\
Boundary Cases & 68.0\% & 52.0\% \\
\bottomrule
\end{tabular}
    }
    }
    \hfill
    \parbox{.48\linewidth}{
        \centering

    \caption{\footnotesize {Dialogue instruction following scores by model and constraint type}}
\label{table:dialogue_scores}
        \resizebox{0.48\columnwidth}{!}{
         \begin{tabular}{lccccr}
\toprule
Model & Overall & Content & Format & Style & Logical \\
\midrule
GPT-4 & \textbf{0.8181} & \textbf{0.4375} & 0.9323 & \textbf{0.9818} & \textbf{0.9779} \\
DeepSeek & 0.7569 & 0.3455 & \textbf{0.9529} & 0.9241 & 0.9031 \\
GPT-3 & 0.7113 & 0.1908 & 0.9083 & 0.9763 & 0.8683 \\
Mixtral & 0.7016 & 0.3568 & 0.3958 & 0.9453 & 0.9557 \\
\bottomrule
\end{tabular}}
        }
\vspace*{-0.6cm}       
\end{table*}

\textbf{Predicate Type Analysis.} DIALEVAL's automated decomposition differs substantially from INFOBENCH's manual approach. DIALEVAL generates finer-grained decompositions for complex instructions (5.4 versus 4.2 predicates per Hard Set instruction), enabling more precise evaluation while requiring robust independence enforcement. Predicate type distributions also differ significantly: DIALEVAL identifies more logical predicates (24\% versus 17\%) and fewer content predicates (38\% versus 45\%). This reflects DIALEVAL's structured classification mechanism that explicitly separates logical reasoning from content requirements, contributing to enhanced performance on complex instructions requiring nuanced assessment of reasoning structures.

\subsection{Dialogue Instruction Following Evaluation}
Applying DIALEVAL to multi-turn dialogues of the BotWars dataset reveals distinct patterns in instruction following capabilities across LLM architectures. 

\textbf{Cross-Model Performance Comparison. }Our evaluation reveals consistent variations in dialogue-based instruction following capabilities across LLM models (Table~\ref{table:dialogue_scores}). GPT-4 achieves the highest overall DIFS score (0.8181), followed by DeepSeek (0.7569), GPT-3 (0.7113), and Mixtral (0.7016). This performance ordering aligns with findings from standard benchmarks such as MMLU~\cite{hendrycks2020measuring}, though with different performance gaps in dialogue contexts.

\textbf{Predicate Type Analysis.} Type-specific evaluation reveals pronounced performance asymmetry across predicate categories. Content predicates present universal challenges: GPT-4 achieves only 0.4375, with other models performing worse (DeepSeek: 0.3455, Mixtral: 0.3568, GPT-3: 0.1908). Notably, content satisfaction consistently measures 40-45\% of style satisfaction across GPT-4, DeepSeek, and Mixtral despite architectural differences, with GPT-3 showing larger disparity (19.5\%). This persistent 55-60\% performance gap across transformer architectures indicates systematic limitations in conditional content generation under multiple simultaneous predicates.

This limitation manifests critically in task-oriented dialogue domains (customer service, information retrieval) where precise content recall must satisfy concurrent stylistic and formatting requirements. The quantifiable performance gap suggests that architectural innovations targeting cross-attention between content representations and predicate tokens may address this limitation.reveals architecture-specific instruction following patterns through type-specific predicate satisfaction analysis.

\begin{figure}[h]
\vspace*{-0.4cm}   
\centering
\resizebox{0.5\linewidth}{!}{
\begin{tikzpicture}
\begin{axis}[
  width=12cm,
  height=8cm,
  xlabel={\textbf{Instruction ID}},
  ylabel={\textbf{Score}},
  xmin=0.5, xmax=7.5,
  ymin=0, ymax=1.1,
  xtick={1,2,3,4,5,6,7},
  ytick={0,0.2,0.4,0.6,0.8,1.0},
  legend style={at={(0.97,0.03)}, anchor=south east, draw=gray!30, fill=white!95, opacity=0.95, rounded corners=2pt},
  legend columns=1,
  ymajorgrids=true,
  xmajorgrids=true,
  grid style={dotted, gray!40},
  axis background/.style={fill=gray!5},
  tick align=outside,
  legend cell align=left,
  every axis plot/.append style={line width=1.5pt},
  title style={font=\bfseries},
  xlabel shift={-3pt},
  ylabel shift={-3pt}
]
\addplot[color=teal!80!black,mark=square*,mark size=3pt] coordinates {
  (1,0.0417) (2,0.9323) (3,0.9635) (4,0.8333) (5,0.9896) (6,1.0) (7,0.9661)
};
\addplot[color=orange!90!black,mark=triangle*,mark size=3.5pt] coordinates {
  (1,0.1309) (2,0.9529) (3,0.8482) (4,0.5602) (5,0.9267) (6,1.0) (7,0.8796)
};
\addplot[color=purple!70!black,mark=o,mark size=3pt] coordinates {
  (1,0.0414) (2,0.9083) (3,0.9527) (4,0.3402) (5,0.8580) (6,1.0) (7,0.8787)
};
\addplot[color=red!70!black,mark=diamond*,mark size=4pt] coordinates {
  (1,0.0417) (2,0.3958) (3,0.9219) (4,0.6719) (5,0.9844) (6,0.9688) (7,0.9271)
};
\legend{\textbf{GPT-4}, \textbf{DeepSeek}, \textbf{GPT-3}, \textbf{Mixtral}}
\end{axis}
\end{tikzpicture}}
\caption{\footnotesize Per-instruction performance. Instruction IDs: (1) Initiate conversation by asking about caller identity, (2) Response length $<$ 30 words (3) Express hesitation for sensitive information, (4) Provide plausible fake information, (5) Maintain conversation flow, (6) Maintain naive persona, (7) Build upon previous dialogue context.}
\label{fig:instruction_comparison}
\vspace*{-0.6cm}   
\end{figure}
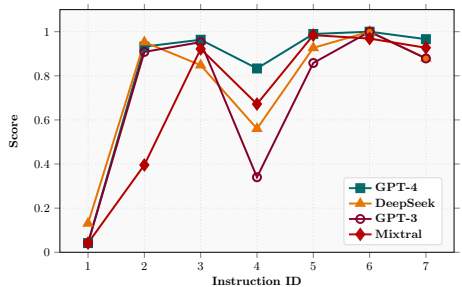

\textbf{Architecture-Constraint Analysis. } Predicate-type performance reveals distinct architectural patterns (Figure~\ref{fig:instruction_comparison}). GPT-4 demonstrates relatively balanced performance across types despite content weaknesses, while Mixtral exhibits distinctive asymmetry: weak format satisfaction (0.3958) paired with strong logical satisfaction (0.9557), consistent with mixture-of-experts architecture where different experts specialize in different capabilities, potentially causing imbalanced routing decisions that favor logical over format processing.

Inter-predicate correlation analysis reveals strong logical-style correlation ($r = 0.82$) vs. weak format-content correlation ($r = 0.13$) across all models, suggesting logical reasoning and style capabilities share underlying computational mechanisms while content and format processing operate through distinct pathways. This supports architectural designs separating content generation from formatting while integrating logical reasoning with style management~\cite{shazeer2017outrageously}. 


\textbf{Instruction-Specific Analysis.} Performance on instruction 1 (initiating conversation by asking caller identity) reveals systematic dialogue initiative limitations persistent across architectures: GPT-3 and GPT-4 achieve virtually identical scores (0.0414 versus 0.0417) despite substantial parameter scaling. This contrasts sharply with instruction 4 (providing plausible fake information), where GPT-4 substantially outperforms GPT-3 (0.8333 versus 0.3402), demonstrating that architectural advancements enhanced contextual reasoning but not dialogue initiative.

Instruction 2 (maintaining response under 30 words) exposes architectural differences in numerical predicate satisfaction: GPT-4 (0.9323) and DeepSeek (0.9529) demonstrate strong performance under exact precision semantics, while Mixtral exhibits marked weakness (0.3958). GPT-4 uniquely maintains high satisfaction across both numerical constraints (instruction 2: 0.9323) and content generation (instruction 4: 0.8333), indicating superior predicate integration—critical for applications requiring precise, concise responses.

Universal success on instruction 6 (maintaining persona, scores $> 0.96$) demonstrates that persona consistency represents a solved capability, aligning with findings that representation predicates are more readily satisfied than generation predicates~\cite{roller2021blenderbot}.

\section{Conclusion}
\label{sec:conclusion}

We presented DIALEVAL, a type-theoretic framework using dual LLM agents to automate instruction evaluation through typed predicate satisfaction. The framework enforces formal atomicity and independence constraints during extraction, then applies type-specific semantics (semantic equivalence for content, exact precision for numerical) that mirror human assessment patterns, eliminating manual annotation. Validation demonstrates substantial improvements over state-of-the-art methods with stronger correlation for complex instructions. Application to multi-turn dialogues reveals universal content predicate challenges (satisfaction 0.19-0.44) despite strong style and logical performance (>0.86), indicating fundamental limitations in conditional content generation under multiple predicates. 
DIALEVAL's type-theoretic approach enables systematic identification of these architectural constraints, providing actionable insights for dialogue system development.

\bibliographystyle{splncs04}
\bibliography{short_bib.bib}

\end{document}